\documentclass[runningheads]{llncs}
\usepackage{graphicx}
\usepackage{cite}
\usepackage[colorlinks=true,linkcolor=blue, citecolor=blue, urlcolor=blue]{hyperref}
\usepackage[misc]{ifsym}
\usepackage{bbm}
\usepackage{url}
\usepackage{amssymb}
\usepackage{amsmath}
\usepackage{multirow}
\usepackage{ifsym}
\usepackage{array}
\usepackage[dvipsnames]{xcolor}

\makeatletter
\newcommand{\printfnsymbol}[1]{%
  \textsuperscript{\@fnsymbol{#1}}%
}
\usepackage[symbol]{footmisc}

\makeatother

\begin{document}
\title{Learning Site-specific Styles for Multi-institutional Unsupervised Cross-modality Domain Adaptation}
\titlerunning{Site-specific Styles for Multi-institutional UDA}

\author{Han Liu\inst{1}\Letter\and
Yubo Fan\inst{1}\and
Zhoubing Xu\inst{2} \and
Benoit M. Dawant\inst{1} \and
Ipek Oguz\inst{1}}
\authorrunning{H. Liu et al.}
\institute{Vanderbilt University, Nashville, TN, USA\and 
Siemens Healthineers, Princeton, NJ, USA\\
\email{han.liu@vanderbilt.edu}}
\maketitle            

% related work
% qualitative results
% discussion, future work

\begin{abstract}
Unsupervised cross-modality domain adaptation is a challenging task in medical image analysis, and it becomes more challenging when source and target domain data are collected from multiple institutions. In this paper, we present our solution to tackle the multi-institutional unsupervised domain adaptation for the crossMoDA 2023 challenge. First, we perform unpaired image translation to translate the source domain images to the target domain, where we design a dynamic network to generate synthetic target domain images with controllable, site-specific styles. Afterwards, we train a segmentation model using the synthetic images and further reduce the domain gap by self-training. Our solution achieved the $1^{st}$ place during both the validation and testing phases of the challenge. The code repository is publicly available at \url{https://github.com/MedICL-VU/crossmoda2023}.

\keywords{MRI \and Vestibular schwannoma \and Cochlea \and Multi-institutional \and Unsupervised domain adaptation \and Style transfer \and Dynamic network}
\end{abstract}

% outline
% key ideas to convey:
% - to tackle domain difference: - harmonize - diversity. we chose diversity
% - we don't want many cyclegans, we also don't want a star-gan. we only need 1x3 mapping
% fig 1. 3D network architecture: dynamic instance norm. instead of 3x3, we learn a 1 to 3 mapping.
% fig 2. our results on generated images. compare to real T2 images.
% fig 3. oversample cases with dynamic codes
% fig 4. segmentation results. good vs. bad.
% Table: our ablation
% implementation details to include:
% - 3D ROI crop. using cochlea segmentation on 2D qs-att
% - preprocessing: T1 image: decrease intensity on tumor, add fake cochlea
% - GAN loss function: edge loss, segmentation losses
% - offline oversample with code augmentation
% - nnunetV2: tumor/cochlea augmentation, only LR flipping
% - train from scratch. self-training (remove unreliable pseudo labels by using connected components)
% - model ensemble: 2 (u-net/res) x 2 (DA, moreDA) x 2 (data/ data+supp)
% our advantage: we do not need site-specific models, we need only a single seg model.
%  can be trained with more T2 images
\newcommand{\xxx}[1]{ \textcolor{red}{#1} }

\section{Introduction}

\begin{figure}[t]
\includegraphics[width=1\columnwidth]{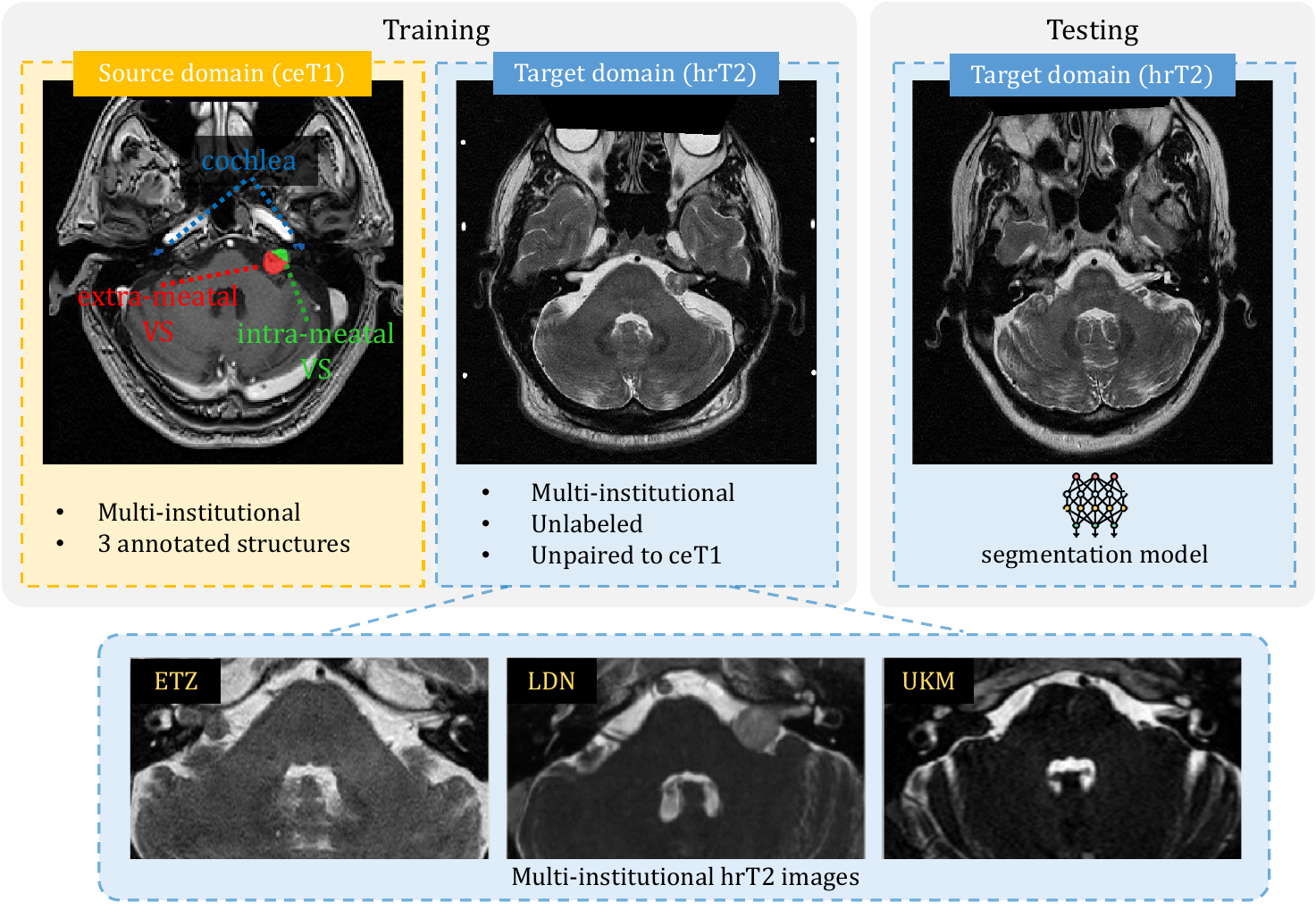}
\centering
\caption{Schematic problem description of the crossMoDA 2023 challenge. The task of this challenge is cross-modality unsupervised domain adaptation (UDA), where source domain and target domain are contrast-enhanced T1 (ceT1) and high-resolution T2 (hrT2), respectively. Note that both source and target domain data are collected from multiple institutions, leading to additional challenges to the UDA tasks, which primarily focus on the inter-domain gap rather than the intra-domain variability.} \label{fig1}
\end{figure}

The crossMoDA challenges\footnote[1]{https://crossmoda-challenge.ml/} \cite{dorent2023crossmoda} aim to tackle the unsupervised cross-modality segmentation of vestibular schwannoma (VS) and cochleae on MRI scans. Specifically, participants are provided with the labeled source domain data, i.e., contrast-enhanced T1-weighted (ceT1) images, and the unlabeled target domain data, i.e., high-resolution T2-weighted (hrT2) images. The goal of this challenge is to train a segmentation model for the target domain hrT2 images. The crossMoDA 2023 extends the previous editions by introducing (1) a sub-segmentation task for the VS (intra- and extra-meatal components)\cite{wijethilake2022boundary} and (2) more heterogeneous data collected from multiple institutions. The schematic problem description of the crossMoDA 2023 is illustrated in Fig. \ref{fig1}. Specifically, the organizers partition the multi-institutional images into 3 sub-datasets, namely ETZ, LDN, and UKM. It can be observed that the hrT2 images from different sub-datasets have significantly different appearances and thus it is critical to ensure the robustness of our segmentation model on the multi-institutional data. 

As the images within the same sub-dataset have relatively consistent styles, we assume that the images within each sub-dataset are collected from the same \textbf{site}. Note that this assumption is not accurate for the UKM sub-dataset as it includes images collected from multiple sites. However, by considering the UKM images as collected from the same site, we will show that our generative model can learn a UKM-specific style that can be used to diversify the styles of our synthetic images. In this paper, our contributions are summarized as follows:
\begin{itemize}
    \item We revisit the top-performing solutions of the previous crossMoDA challenges and analyze the factors contributing to their success.
    \item To addresses the intra-domain variability in multi-institutional UDA, we propose a dynamic network to generate synthetic images with controllable, site-specific styles, which are used to train the downstream segmentation model for improved robustness and generalizability.
    \item Our proposed method achieves the $1^{st}$ place during both the validation and testing phases of the crossMoDA 2023 challenge.
\end{itemize}

\section{Related Works}
\subsection{CrossMoDA Challenges: 2021-2023}
While numerous domain adaptation techniques have been proposed for image segmentation, most of these techniques have only been validated either on private datasets or on small public datasets, and mostly addressed single-class segmentation tasks. The crossMoDA challenge\cite{dorent2023crossmoda} introduced the first large and multi-class dataset for cross-modality domain adaptation for medical image segmentation. In the 2021 edition, source and target domain data were collected from a single scanner and the participants were asked to segment the cochleae and the whole VS in hrT2 images, i.e., a 2-class segmentation task. With the same task, the 2022 edition included additional data from another scanner for both source and target domain datasets, making the domain adaptation task more challenging by introducing intra-domain variability. The 2023 edition further enlarged the datasets by including multi-institutional, heterogeneous data for both domains and introduced a sub-segmentation for the VS (intra- and extra-meatal components), leading to a 3-class segmentation task with significant intra-domain variability.

\begin{figure}[t]
\includegraphics[width=1\columnwidth]{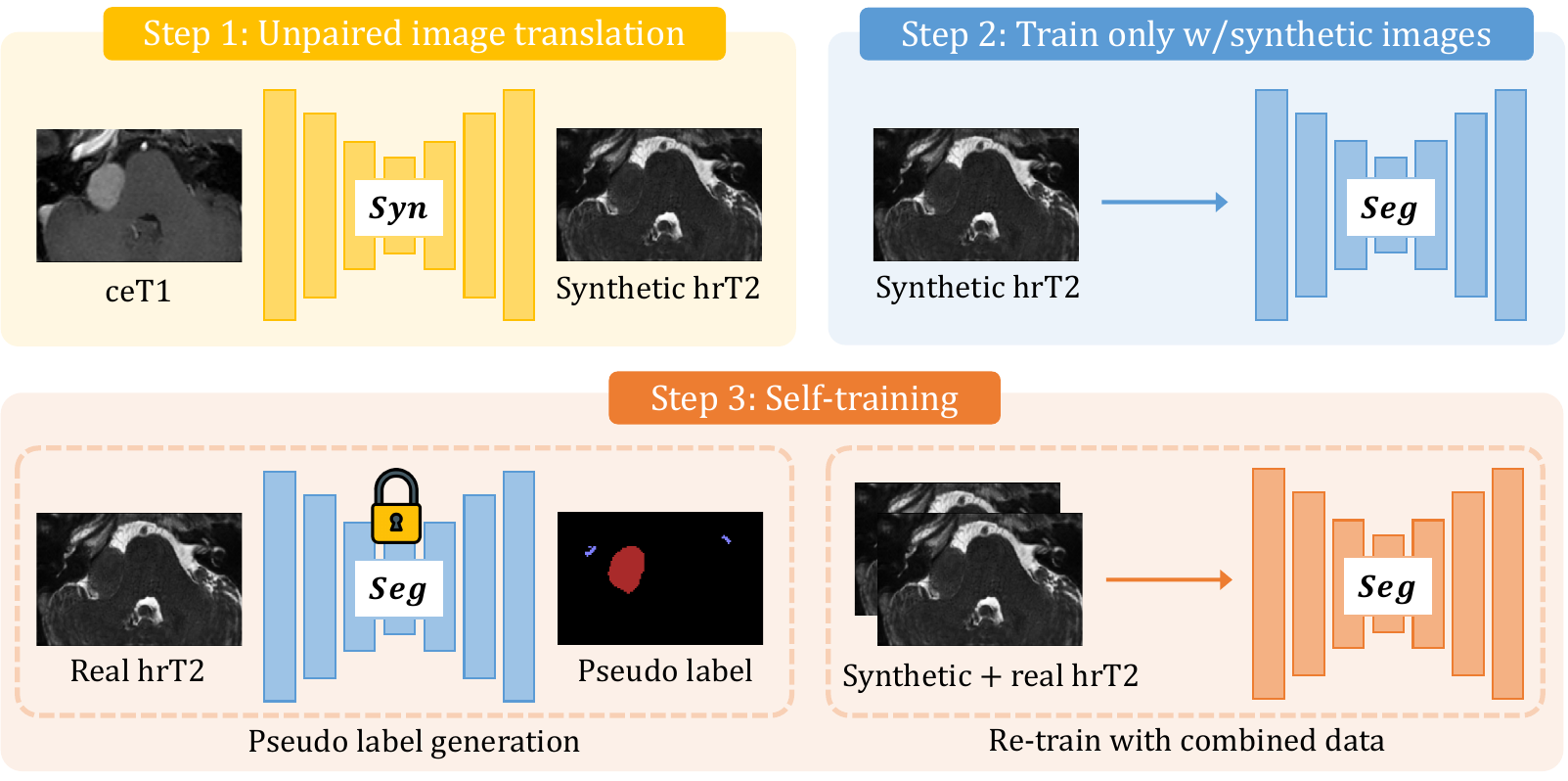}
\centering
\caption{The training strategy of the image-level domain alignment approaches for UDA.} \label{fig2}
\end{figure} 

\subsection{Top Solutions in crossMoDA 2021 and 2022}
\label{3step}
The top solutions in the 2021 and 2022 editions are mainly based on the \textbf{\textit{image-level domain alignment}} approach. As illustrated in Fig. \ref{fig2}, it typically consists of three steps. In step 1, unpaired image translation is used to translate ceT1 images to synthetic hrT2 images. The most commonly used techniques include cycleGAN\cite{zhu2017unpaired}, CUT\cite{park2020contrastive}, QS-Attn\cite{hu2022qs} with either 2D or 3D backbones. In step 2, the synthetic hrT2 images and the associated ceT1 labels are used to train a segmentation model. In step 3, to further reduce the domain gap between synthetic and real hrT2, the unlabeled real hrT2 are used to train the segmentation model via self-training. Specifically, the network trained in step 2 is used to firstly generate the pseudo labels on the real hrT2 images. Then the synthetic and real hrT2 images are combined to re-train a segmentation network. This self-training process can be repeated iteratively by using the most updated pseudo labels generated by the network trained at the previous iteration.

Based on the image-level domain alignment strategy, the top teams have proposed a variety of techniques to further improve the performance. In the 2021 edition, the 1$^{st}$ place team\cite{shin2022cosmos} proposed to add segmentation decoders to the generators of the 2D cycleGAN to better synthesize the VS and the cochlea. Additionally, they visually inspected the pseudo labels to select the most reliable ones for self-training. The 2$^{nd}$ place team proposed PAST\cite{dong2021unsupervised}, where 2D NICE-GAN\cite{chen2020reusing} was used for image synthesis and self-training with pixel-level pseudo label filtering was used for segmentation. The 3$^{rd}$ place team\cite{choi2021using} used the CUT model for image synthesis and proposed an offline data augmentation technique to simulate the heterogeneous signal intensity of VS. In the 2022 edition, the 1$^{st}$ place team built upon the PAST algorithm and added extra segmentation heads for NICE-GAN. Moreover, to address the intra-domain variability, they trained separate segmentation models for different sites and structures. The 2$^{nd}$ place team\cite{kang2023multi} proposed to improve the image synthesis via multi-view image translation, where the cycleGAN and the QS-Attn were used in parallel. The 3$^{rd}$ place team\cite{salle2023cross} proposed to improve the generalizability of the segmentation model by generating diverse appearances of VS via SinGAN\cite{shaham2019singan}.

In summary, the top solutions in 2021 and 2022 editions demonstrated three promising directions to improve the image-level domain alignment: (1) better synthetic hrT2 images in step 1, (2) higher-quality pseudo labels for self-training in step 3, and (3) local intensity augmentation for VS in step 2 and 3. 

\section{Methods}
Motivated by the previous works\cite{liu2021unsupervised,li2021unsupervised,shin2022cosmos,dong2021unsupervised,han2022unsupervised,zhao2023ms}, we propose to tackle the UDA problem by reducing the domain gap at the image-level, and follow the 3-step strategy as presented in Sec. \ref{3step}. Since the quality of synthetic hrT2 images is critical to the performance of the downstream segmentation task, our key innovations are mainly focused on the step 1, i.e., unpaired image translation. To address the intra-domain variability, we propose to generate synthetic hrT2 images with \textbf{site-specific styles}, which are then used to train the segmentation model for improved robustness to various hrT2 styles. The details of our novel techniques for image translation are provided as follows.

\begin{figure}[t]
\includegraphics[width=1\columnwidth]{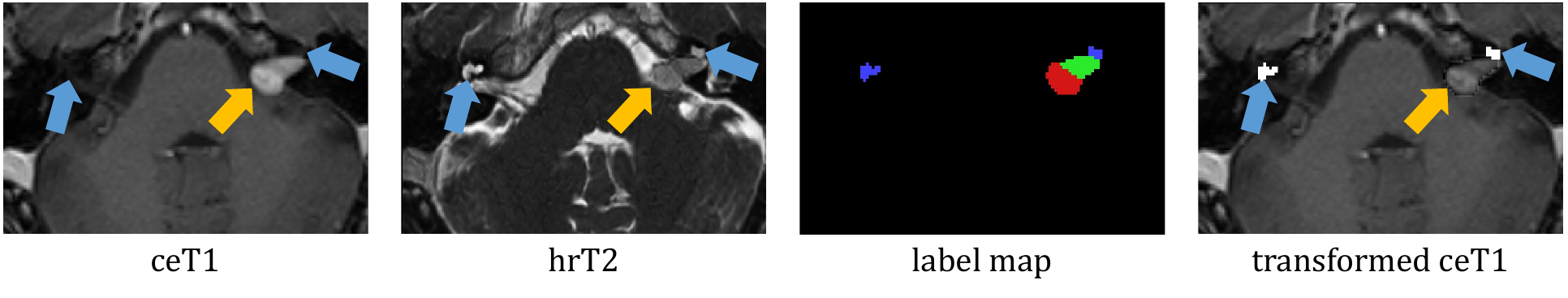}
\centering
\caption{Illustration of our proposed label-assisted intensity transformation. The VS (yellow arrow) and cochleae (blue arrows) have opposite intensity profiles in ceT1 and hrT2 images.} \label{fig3}
\end{figure}

\subsection{Label-assisted Intensity Transformation}
The VS and the cochleae have significantly different intensity profiles in ceT1 and hrT2. As shown in Fig. \ref{fig3}, the cochleae have weak signals and the VS has strong signals in ceT1 images, but the opposite is true in hrT2. Our preliminary experiment shows that the synthesis network with the original ceT1 as input may fail to capture the appearance difference of these structures between the two modalities. To address this problem, we propose to transform the intensity profiles of VS and cochlea in ceT1 images before feeding them to the synthesis network. After we perform regular preprocessing steps, which include rescaling to [-1, 1] range (see Sec. \ref{preprocessing}), we replace the intensity values of the cochleae by 1.0, i.e., the maximum value of the preprocessed image. In addition, we decrease the intensity values of the VS by $\mu_{VS}+0.5$, where $\mu_{VS}$ is the mean intensity of the VS. With the transformed ceT1, we reduce the appearance difference of the VS and cochleae across two modalities before performing the synthesis task.

\begin{figure}[!b]
\includegraphics[width=1\columnwidth]{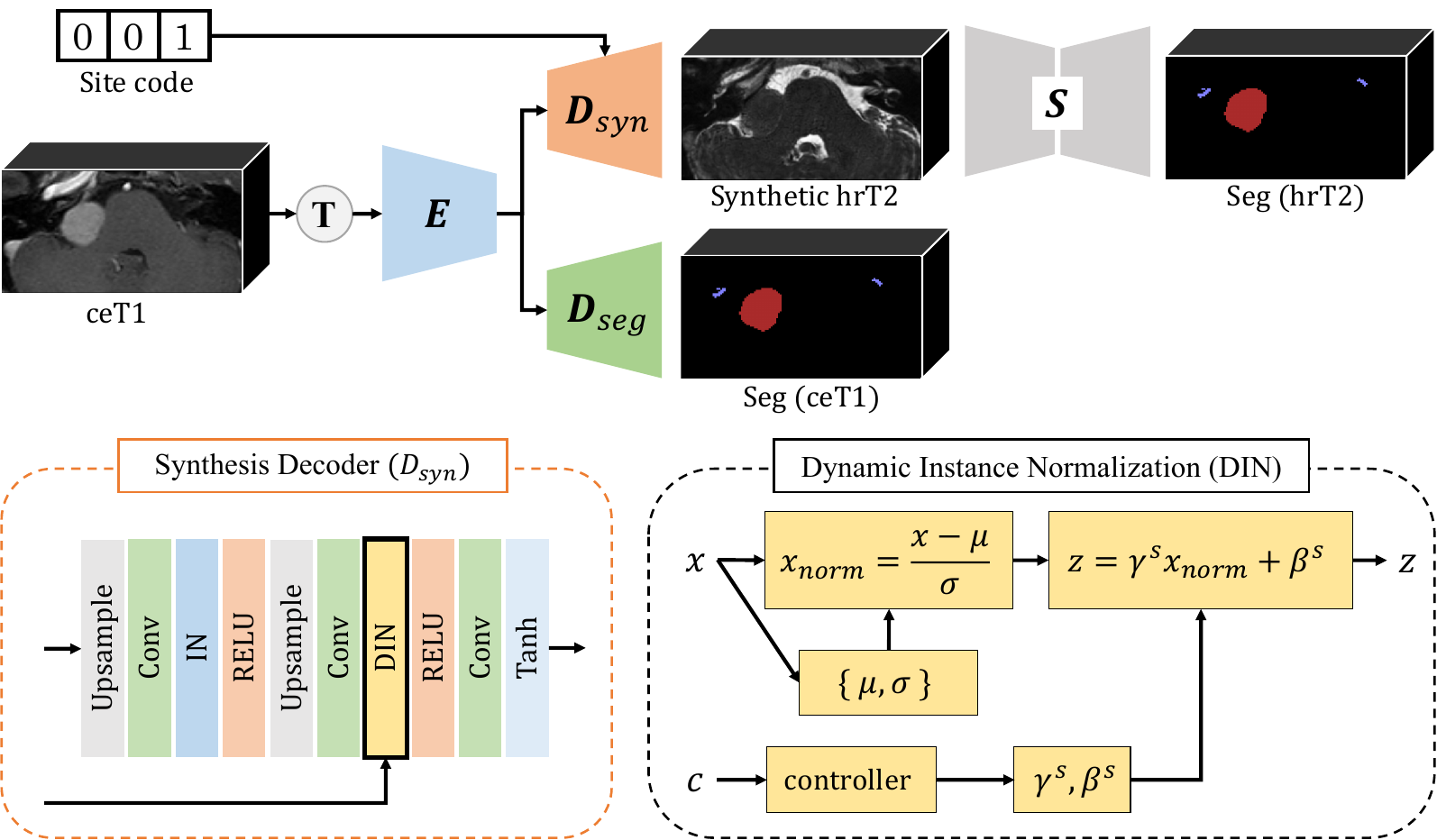}
\centering
\caption{Illustration of our dynamic generator used for the unpaired image translation. T is the label-assisted intensity transformation. Given a site code (a one-hot vector), our dynamic network is trained to generate site-specific affine parameters for the last instance normalization layer, which is then used to control the output hrT2 styles.} \label{fig4}
\end{figure}

\subsection{Anatomy-aware Image Synthesis}
We adopt the QS-Attn \cite{hu2022qs} and extend it to 3D for volumetric unpaired image translation. 3D QS-Attn is used because (1) compared to 2D networks, 3D networks can generate synthetic images with better slice-to-slice continuity by exploiting the intra-slice information, and (2) compared to CycleGAN\cite{zhu2017unpaired}, QS-Attn is less memory-intensive and thus more suitable for 3D networks. 

As shown in Fig. \ref{fig4}, we propose to improve the image synthesis by making the generator focus more on the anatomical structures in the downstream segmentation task, i.e., the VS and the cochleae. To this end, we add an extra segmentation decoder $D_{seg}$ to the generator such that our generator learns to synthesize hrT2 images and segment these structures jointly. As demonstrated in \cite{shin2022cosmos}, this multi-task learning paradigm can help better preserve the shape of the structures-of-interest (SOI) in the synthetic images. Moreover, we employ another segmentation network $S$ to segment SOI from the synthetic hrT2 images, further encouraging the generated SOI to have semantically meaningful boundaries.

\subsection{Site-specific Styles}
\begin{figure}[!ht]
\includegraphics[width=1\columnwidth]{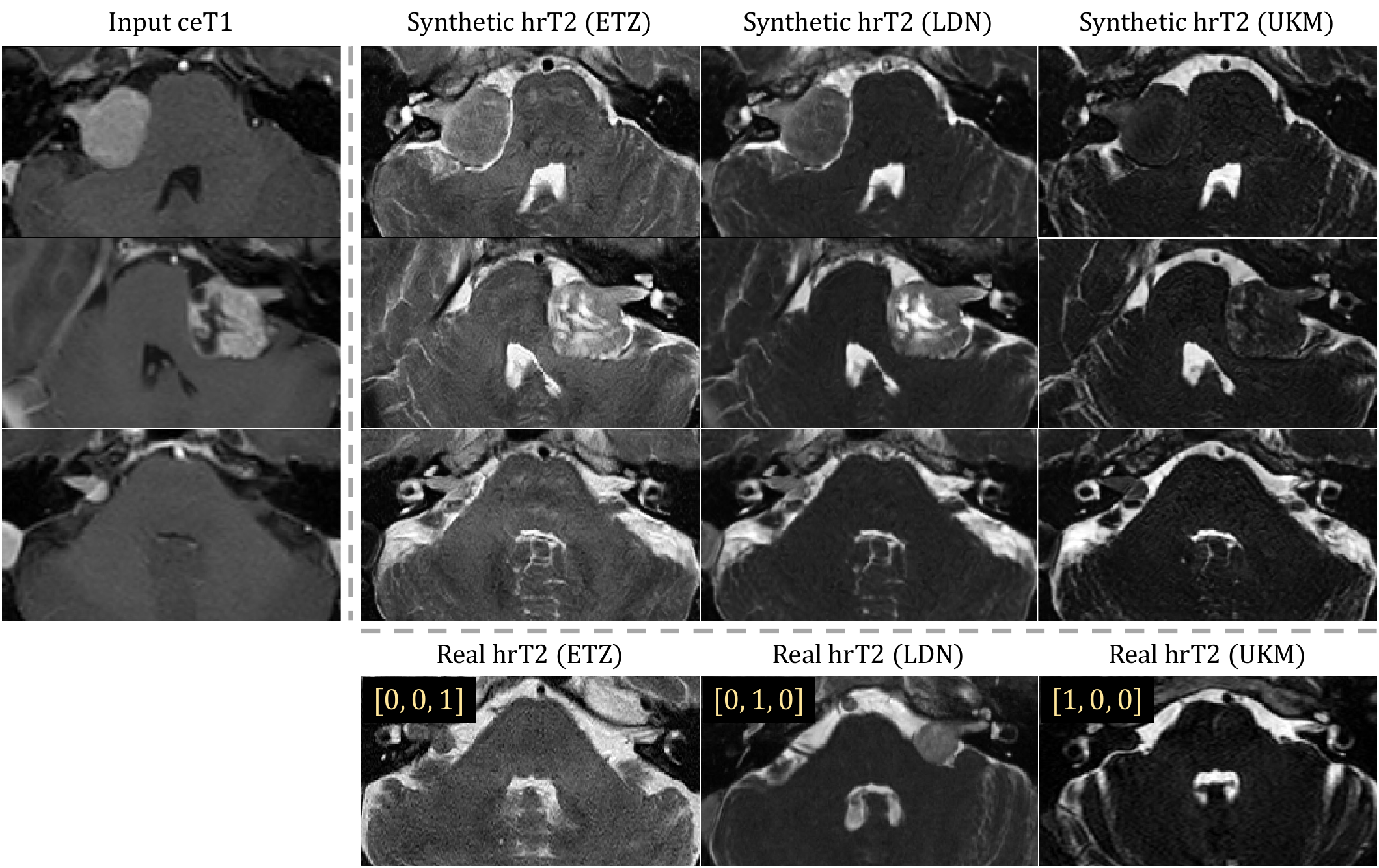}
\centering
\caption{Synthetic hrT2 images with site-specific styles. In top three rows, each row displays a representative ceT1 image being transformed to hrT2 with different site-specific styles. The bottom row displays real hrT2 images from three different sites, which are used as references for style comparison. Each column corresponds to the same site-specific style and the associated site code is shown on the top left corner at the bottom row.} \label{fig5}
\end{figure}

To ensure the robustness to different hrT2 styles, we propose to generate the synthetic hrT2 images with site-specific styles to train the segmentation model. Inspired by \cite{liu2022moddrop++}, we propose to modify the synthesis decoder to a dynamic network, where the style of the output hrT2 image is conditioned on a given site prior. Specifically, we replace the last instance normalization (IN) layer of the synthesis decoder by a dynamic instance normalization (DIN) layer. This is motivated by previous studies\cite{dumoulin2017a,huang2017arbitrary,karras2019style} where the IN layers are shown to effectively control the \textbf{styles} of images. We encode the site condition as a one-hot vector $c$, which is passed to a controller (a 3D convolutional layer with a kernel size of $1\times1\times1$) to generate site-specific affine parameters $\gamma^{s}$ and $\beta^{s}$ for IN. Therefore, we can train a single unified synthesis network on all hrT2 images with a controllable output style, as shown in Fig. \ref{fig5}. 

\begin{figure}[!ht]
\includegraphics[width=1\columnwidth]{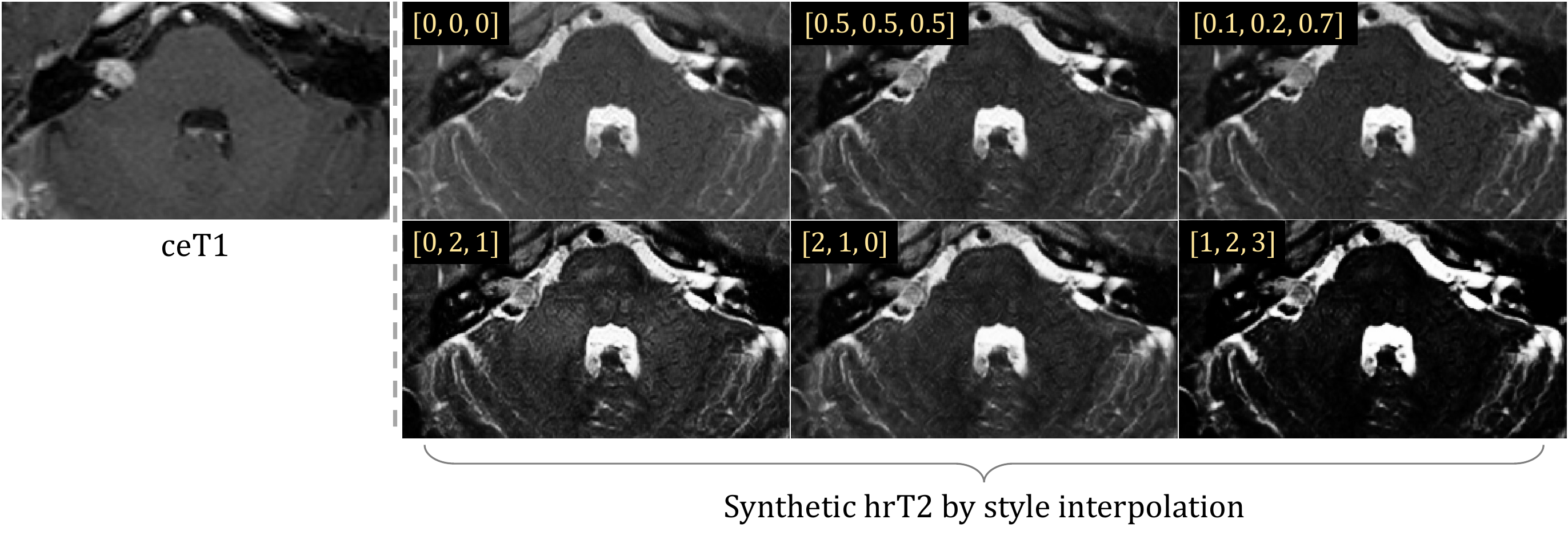}
\centering
\caption{Examples of hrT2 styles generated by style interpolation. During inference, arbitrary site codes can be used as the site condition to generate unseen hrT2 styles. In each example, the site code is shown at the top left corner 
.} \label{fig6}
\end{figure} 

\subsection{Oversampling Hard Samples by Style Interpolation}\label{sec3.4}
Based on the segmentation results from the validation set, we observe that the VS with either (1) tiny/no extra-meatal components, or (2) large extra-meatal components with heterogeneous appearance, are more challenging to segment. We refer to these cases as \textit{hard samples}. We find that such hard samples are indeed under-represented in the source domain dataset and their associated synthetic hrT2 may need to be oversampled for balanced training. In practice, we select the hard samples based on the aforementioned two rules with the help of source domain labels. Inspired by style interpolation \cite{dumoulin2017a,huang2017arbitrary,hu2023map}, we propose to generate more diverse hrT2 styles for oversampling by feeding the controller with unseen site codes. As shown in Fig. \ref{fig6}, we oversample each hard sample by translating the same ceT1 image into a variety of unseen hrT2 styles, further enriching the diversity of our synthetic dataset.

\subsection{Implementation Details}
\subsubsection{Preprocessing}\label{preprocessing} All MR scans are set to the RAI orientation, resampled to the median voxel size of the dataset, i.e., $0.41\times0.41\times1$ mm$^{3}$, and further cropped into $256\times144\times32$ based on the positions of the cochleae, which are computed by a segmentation network additionally trained on real hrT2 images with pseudo-labeled cochleae. The cropped volumes are used for all the synthesis and segmentation tasks. For image synthesis, we normalize both ceT1 and hrT2 images using Z-score normalization, clip the intensity values to the [0, 99.9$^{th}$] percentile, and rescale the values to [-1, 1]. 

\subsubsection{Synthesis} The backbone of our dynamic generator is a 3D 9-block ResNet. Due to the limit of GPU memory, the input is a 3D patch with a size of $256\times144\times8$ randomly cropped from the preprocessed image. We use overlapping sliding windows for inference. During training, we apply on-the-fly data augmentation including random contrast adjustment ($p=0.4$ for ceT1 and $p=0.1$ for hrT2; smaller $p$ for hrT2 to preserve the site-specific style) and randomly flipping on the LR direction ($p=0.5$). The loss function for image synthesis is expressed as: $L_{G}= L_{QS}+\lambda_{1}L_{seg}^{ceT1}+\lambda_{2}L_{seg}^{hrT2}+\lambda_{3}L_{edge}$, where $L_{QS}=L_{adv}+L_{con}^{ceT1}+L_{con}^{hrT2}$ is the default loss function of QS-Attn. $L_{seg}^{ceT1}$ and $L_{seg}^{hrT2}$ are the segmentation losses for $D_{seg}$ and $S$, respectively. We also adopt an edge loss $L_{edge}$\cite{yu2019ea,fan2023temporal,liu2023evaluation}  to encourage the edge consistency between the input and the output so that the texture within the VS and the cochlea boundary can be well preserved. We use $\lambda_{1}=0.5$, $\lambda_{2}=0.5$ and $\lambda_{3}=1$. We train the network for 400 epochs with a learning rate of $2e-4$ and another 400 epochs with linear decay policy. For other hyperparameters, we use the default settings of the QS-Attn. 

\subsubsection{Segmentation} We use the nnU-Net V2\cite{isensee2021nnu} with 3D fullres configuration for all our segmentation tasks. We build upon the default nnUNetTrainer and make the following modifications. First, we only enable random flipping along LR direction. Second, we introduce two local intensity augmentation functions to only augment the intensity values of the VS and the cochlea. Specifically, we randomly multiply the VS intensity with $u\sim U(1.2, 2)$. In addition, we randomly reduce the cochleae intensity by $v\sim U(0.5, 1)$, since previous study indicates that the cochleae ipsilateral to VS may have weaker signals in hrT2\cite{cass2023automated}. We follow \cite{liu2022enhancing} to train segmentation models and perform two rounds of self-training. Previous studies suggest that image-level pseudo label filtering can be incorporated into self-training to avoid performance degradation caused by unreliable pseudo labels\cite{huang2023revisiting,yang2022st++,liu2023cosst}. Therefore, we remove the real hrT2 images with unreliable pseudo labels from our training set throughout the self-training process, where the pseudo labels with no tumor prediction or with multiple tumor components on both sides are considered unreliable. Note that the reliability of pseudo labels can be determined by connected component analysis and thus the entire process is fully automatic. Lastly, we use model ensemble by averaging the predictions from 11 models to further boost the performance. These models include 3 standard nnU-Net models trained with different seeds and 8 customized nnU-Net models with the following configurations: 2 different backbone architectures (U-Net or ResU-Net) $\times$ 2 different augmentation strategies (strong or weak local intensity augmentation for VS and cochleae) $\times$ 2 different sets of unseen site codes for style interpolation.

\section{Experiments and Results}
We use the dataset\footnote[1
]{https://www.synapse.org/\#!Synapse:syn51317912} provided by the crossMoDA 2023 challenge\cite{kujawa2022deep,wijethilake2022boundary}. Dice score and average symmetric surface distance (ASSD) for extra-meatal VS, intra-meatal VS, and cochleae, as well as the boundary ASSD (denoted as `bound') are used for quantitative evaluation. 

\begin{table}[t]
\centering
\caption{Quantitative results during the \underline{validation} phase (96 cases). \textbf{Bold} represents the best scores. The three rows at the bottom are our ablation studies. ST: self-training. Tr: modified nnUNetTrainer. OS: oversampling}
\label{tab1}
\begin{tabular}{|>{\centering\arraybackslash}m{2.5cm}|>{\centering\arraybackslash}m{1.2cm}|>{\centering\arraybackslash}m{1.2cm}|>{\centering\arraybackslash}m{1.2cm}|>{\centering\arraybackslash}m{1.2cm}|>{\centering\arraybackslash}m{1.2cm}|>{\centering\arraybackslash}m{1.2cm}|>{\centering\arraybackslash}m{1.2cm}|}
\hline 
\multirow{2}{*}{Method} & \multicolumn{3}{c|}{Dice$\uparrow$ (\%)} & \multicolumn{4}{c|}{ASSD$\downarrow$ (mm)}\\ 
\cline{2-8} 
  & extra & intra & cochlea & extra & intra & cochlea & bound\\ 
\hline
Ours (ensemble) &\textbf{85.75} &\textbf{74.36} &84.07 &\textbf{0.45} &\textbf{0.44} &0.20 &\textbf{0.51}\\
\hline
Ours (single)   &85.08 &73.34 &84.44 &0.48 &0.45 &0.20 &0.53\\
\hline\hline
Team A &83.63 &70.57 &83.55 &0.50 &0.59 &0.23 &4.76 \\
\hline
Team B &72.75 &56.94 &\textbf{86.66} &17.62 &16.80 &\textbf{0.18} &32.79 \\
\hline
Team C &81.32 &59.79 &83.56 &8.57 &7.67 &0.22 &20.78 \\
\hline\hline
w/o ST &84.06 &71.42 &82.78 &0.50 &0.59 &0.21 &0.54 \\
\hline
w/o (ST, Tr) &81.74 &68.92 &82.81 &8.58 &7.93 &0.53 &8.69 \\
\hline
w/o (ST, Tr, OS) &79.09 &64.15 &81.94 &8.68 &11.80 &0.56 &12.86\\
\hline
\end{tabular}
\end{table}
\begin{table}[!t]
\centering
\caption{Quantitative results during the \underline{testing} phase (341 cases).}
\label{tab2}
\begin{tabular}{|>{\centering\arraybackslash}m{2.5cm}|>{\centering\arraybackslash}m{1.2cm}|>{\centering\arraybackslash}m{1.2cm}|>{\centering\arraybackslash}m{1.2cm}|>{\centering\arraybackslash}m{1.2cm}|>{\centering\arraybackslash}m{1.2cm}|>{\centering\arraybackslash}m{1.2cm}|>{\centering\arraybackslash}m{1.2cm}|}
\hline 
\multirow{2}{*}{Method} & \multicolumn{3}{c|}{Dice$\uparrow$ (\%)} & \multicolumn{4}{c|}{ASSD$\downarrow$ (mm)}\\ 
\cline{2-8} 
  & extra & intra & cochlea & extra & intra & cochlea & bound\\ 
\hline
Ours &\textbf{84.9} &\textbf{72.8} &83.6 &\textbf{0.452} &\textbf{0.496} &\textbf{0.201} &\textbf{0.675}\\
\hline
Team A &80.8 &69.9 &\textbf{84.4} &0.593 &0.581 &0.207 &1.985 \\
\hline
Team B &78.6 &60.7 &84.3 &6.552 &9.711 &0.246 &18.575 \\
\hline
Team C &78.4 &64.6 &81.4 &1.625 &4.036 &1.316 &9.953 \\
\hline
Team D &63.7 &55.8 &75.0 &20.806 &27.814 &12.776 &24.089 \\
\hline
Team E &67.6 &56.3 &76.7 &13.874 &18.607 &11.026 &35.848 \\
\hline
\end{tabular}
\end{table}

\subsection{Quantitative Results} In Table \ref{tab1}, we report the evaluation metrics on the validation leaderboard. Our method (a single model) achieves the $1^{st}$ place on validation leaderboard and model ensembling can slightly improve the performance. Moreover, we perform ablation studies on the validation set to investigate the effectiveness of self-training, our modified nnUNetTrainer, and the oversampling strategy. The results show that each component can effectively improve the segmentation performance. As shown in Table \ref{tab2}, during the testing phase, our method outperforms other methods in all evaluation metrics except the Dice score of cochleae. We note that during both validation and testing phases our method achieves significantly smaller boundary ASSD, i.e., the distance between the intra-meatal and extra-meatal boundary. This demonstrates its superiority in identifying the anatomical separation between the two tumor components.

\begin{figure}[htbp!]
\includegraphics[width=1\columnwidth]{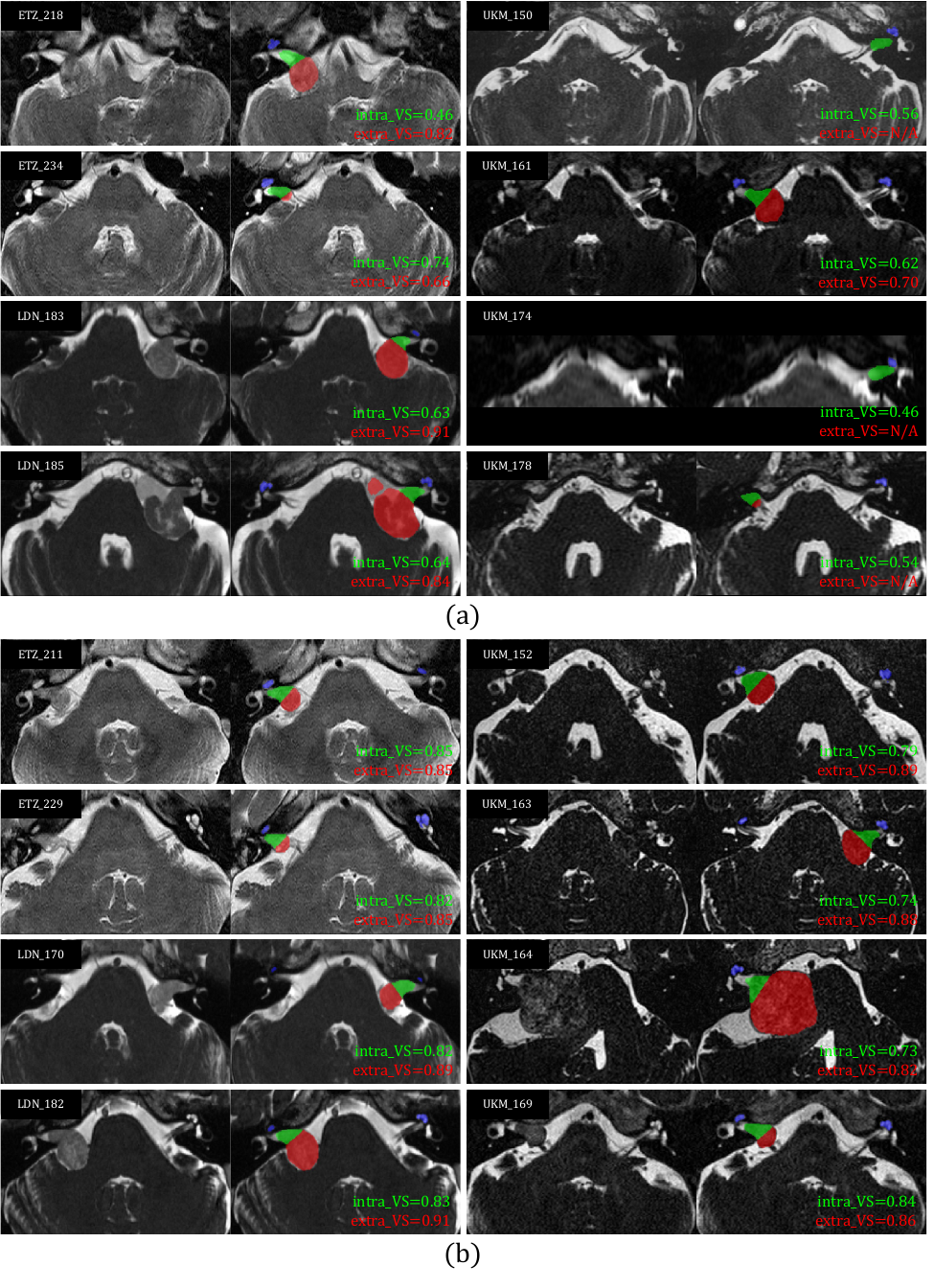}
\centering
\caption{Qualitative results of the representative cases from the validation sets. (a) Unsatisfactory segmentation results. (b) Satisfactory segmentation results. Dice scores of the intra- and extra-meatal VS are displayed.} \label{fig7}
\end{figure} 

\subsection{Qualitative Results} Representative examples of results obtained with images in the validation set are shown in Fig. \ref{fig7}. In Fig. \ref{fig7} (a), we can observe that even with our oversampling technique (Sec. \ref{sec3.4}), the segmentation results on some hard cases remain unsatisfactory. For example, UKM\_150 and LDN\_185 include VS with tiny/no extra-meatal components and VS with large extra-meatal components and heterogeneous textures, respectively. Moreover, the field of view and the image quality may also have a negative impact on the segmentation performance, e.g., UKM\_174. In Fig. \ref{fig7} (b), even though the hrT2 images may have very different styles, our model can produce good segmentation results for the VS whose shapes are common in the training set, indicating that generating site-specific styles is a promising way to improve model robustness for multi-institutional data. 

\section{Discussion and Conclusion}
In this paper, we have presented our solution for the crossMoDA 2023 challenge to tackle the multi-institutional UDA problem. Specifically, we have generated synthetic hrT2 images with site-specific styles to improve the robustness of the segmentation model. The results obtained during both the validation and testing phases show that our method has achieved superior performance against other competitors. Notably, the boundary ASSD achieved by our method is much smaller than the ones achieved by other methods. This suggests that our method is more reliable than other approaches for the follow-up clinical analyses, for which the clear separation between intra- and extra-components is crucial. For instance, the size and volume features extracted from the extra-meatal VS are considered as the most sensitive radiomic features for the evaluation of VS growth\cite{baccianella2009evaluation}.

Though our solution has achieved promising performance, we believe there are several interesting directions to further improve our method. First, by generating site-specific styles, we assume that the images in each sub-dataset are collected from the same site and have relatively consistent appearances. However, this assumption is not strictly accurate for the UKM sub-dataset, where the images are collected from multiple sites and scanners. Indeed, we find that the images in the UKM sub-dataset may have significantly different appearances, which cannot be simply represented by a single site-specific style. Therefore, an interesting direction for future studies is to transform the \textbf{site}-specific style to the \textbf{image}-specific style, i.e., the generated style is conditioned on a reference real hrT2 image. Second, though we can produce some synthetic styles by feeding the dynamic generator with unseen site codes (Sec. \ref{sec3.4}), the generated styles and the associated codes do not have strong correspondence and thus our style interpolation process is not explainable. The underlying reason may be that the dynamic generator is optimized to learn only 3 discrete site-specific styles, leading to a discontinuous latent space of styles. In the future, we will explore some regularization techniques to make the latent space more continuous. This would permit to not only generate site-specific styles, but also more diverse and explainable synthetic styles via style interpolation.

\section{Acknowledgements}
This work was supported in part by the National Science Foundation grant 2220401 and the National Institutes of Health grant T32EB021937.

\bibliographystyle{splncs04}
\bibliography{references.bib}
\end{document}